%% file: main.tex
\documentclass[11pt,a4paper]{article}
\usepackage[nohyperref]{naaclhlt2019}
\usepackage{times}
\usepackage{latexsym}

\usepackage{url}

\aclfinalcopy % Uncomment this line for the final submission
\def\aclpaperid{1614} %  Enter the acl Paper ID here

\usepackage{graphicx}
\usepackage{subfig}
\usepackage{bm}
\usepackage{amssymb}
\usepackage{amsmath}
\usepackage{booktabs}
\usepackage{xcolor}
\usepackage{subfig}
\usepackage{multirow}
\usepackage{booktabs}
\usepackage{verbatim}

\DeclareMathOperator*{\argmax}{arg\,max}
\DeclareMathOperator*{\softmax}{softmax}
\DeclareMathOperator*{\uniform}{Uniform}

\title{Differentiable Sampling with Flexible Reference Word Order\\for Neural Machine Translation}

\author{Weijia Xu \\
	University of Maryland \\
	{\tt \href{mailto:weijia@cs.umd.edu}{weijia@cs.umd.edu}} \\\And
	Xing Niu \\
	University of Maryland \\
	{\tt \href{mailto:xingniu@cs.umd.edu}{xingniu@cs.umd.edu}} \\\And
	Marine Carpuat \\
	University of Maryland \\
	{\tt \href{mailto:marine@cs.umd.edu}{marine@cs.umd.edu}} \\}

\date{}

\begin{document}
\maketitle
\begin{abstract}
Despite some empirical success at correcting exposure bias in machine translation, scheduled sampling algorithms suffer from a major drawback: they incorrectly assume that words in the reference translations and in sampled sequences are aligned at each time step. Our new differentiable sampling algorithm addresses this issue by optimizing the probability that the reference can be aligned with the sampled output, based on a soft alignment predicted by the model itself. As a result, the output distribution at each time step is evaluated with respect to the whole predicted sequence.
Experiments on IWSLT translation tasks show that our approach improves BLEU compared to maximum likelihood and scheduled sampling baselines. In addition, our approach is simpler to train with no need for sampling schedule and yields models that achieve larger improvements with smaller beam sizes.\footnote{The code is available at \url{https://github.com/Izecson/saml-nmt}}
\end{abstract}

\section{Introduction}

Neural machine translation (NMT) models are typically trained to maximize the likelihood of reference translations~\citep{SutskeverVL14,BahdanauCB15}. While simple and effective, this objective suffers from the exposure bias problem~\citep{RanzatoCAZ15}: the model is only exposed to reference target sequences during training, but has to rely on its own predictions at inference. As a result, errors can accumulate along the generated sequence at inference time.  

 This is a well-known issue in sequential decision making~\citep[i.a.]{LangfordZ05,CohenCarvalho2005,KaariainenLangford2006} and
it has been addressed in past work by incorporating the previous decoding choices into the training scheme, using imitation learning~\citep{DaumeLM09,RossGB11,BengioVJS15,LeblondAOL18} and reinforcement learning~\citep{RanzatoCAZ15,BahdanauBXGLPCB16} techniques. In this paper, we focus on a simple and computationally inexpensive family of approaches, known as Data as Demonstrator~\citep{VenkatramanHB15} and scheduled sampling~\citep{BengioVJS15,GoyalDB17}. The algorithms use a stochastic mixture of the reference words and model predictions with an annealing schedule controlling the mixture probability. Despite their empirical success in various sequence prediction tasks, they are based on an assumption that does not hold for machine translation: they assume that words in the reference translations and in sampled sequences are aligned at each time step, which results in weak and sometimes misleading training signals.

In this paper, we introduce a differentiable sampling algorithm that exposes machine translation models to their own predictions during training, and allows for differences in word order when comparing model outputs with  reference translations. We compute the probability that the reference can be aligned with the sampled output using a soft alignment predicted based on the model states, so that the model will not be punished too severely for producing hypotheses that deviate from the reference, as long as the hypotheses can still be aligned with the reference.

Experiments on three IWSLT tasks (German-English, English-German and Vietnamese-English) show that our approach significantly improves BLEU compared to both maximum likelihood and scheduled sampling baselines. We also provide evidence that our approach addresses exposure bias by decoding with varying beam sizes, and show  that our approach is simpler to train than scheduled sampling as it requires no annealing schedule.

\input{approach}
\input{experiments}
\input{related}

\section{Conclusion}

We introduced a differentiable sampling algorithm which exposes a sequence-to-sequence model to its own predictions during training and compares them to reference sequences flexibly to back-propagate reliable error signals. By soft aligning reference and sampled sequences, our approach consistently improves BLEU over maximum likelihood and scheduled sampling baselines on three IWSLT tasks, with larger improvements for greedy search and smaller beam sizes. Our approach is also simple to train, as it does not require any sampling schedule.

\section*{Acknowledgments}

We thank  the anonymous reviewers, Amr Sharaf, Naomi Feldman, Hal Daum\'e III and the CLIP lab at UMD for helpful comments. This research is supported in part by an Amazon Web Services Machine Learning Research Award and by the Office of the Director of National Intelligence (ODNI), Intelligence Advanced Research Projects Activity (IARPA), via contract \#FA8650-17-C-9117. The views and conclusions contained herein are those of the authors and should not be interpreted as necessarily representing the official policies, either expressed or implied, of ODNI, IARPA, or the U.S. Government. The U.S. Government is authorized to reproduce and distribute reprints for governmental purposes notwithstanding any copyright annotation therein.

\bibliography{naaclhlt2019}
\bibliographystyle{acl_natbib}

\end{document}

%% file: approach.tex
\section{Approach}
\begin{figure*}
    \centering
    \subfloat[Scheduled Sampling Objective]{{\includegraphics[width=.55\linewidth]{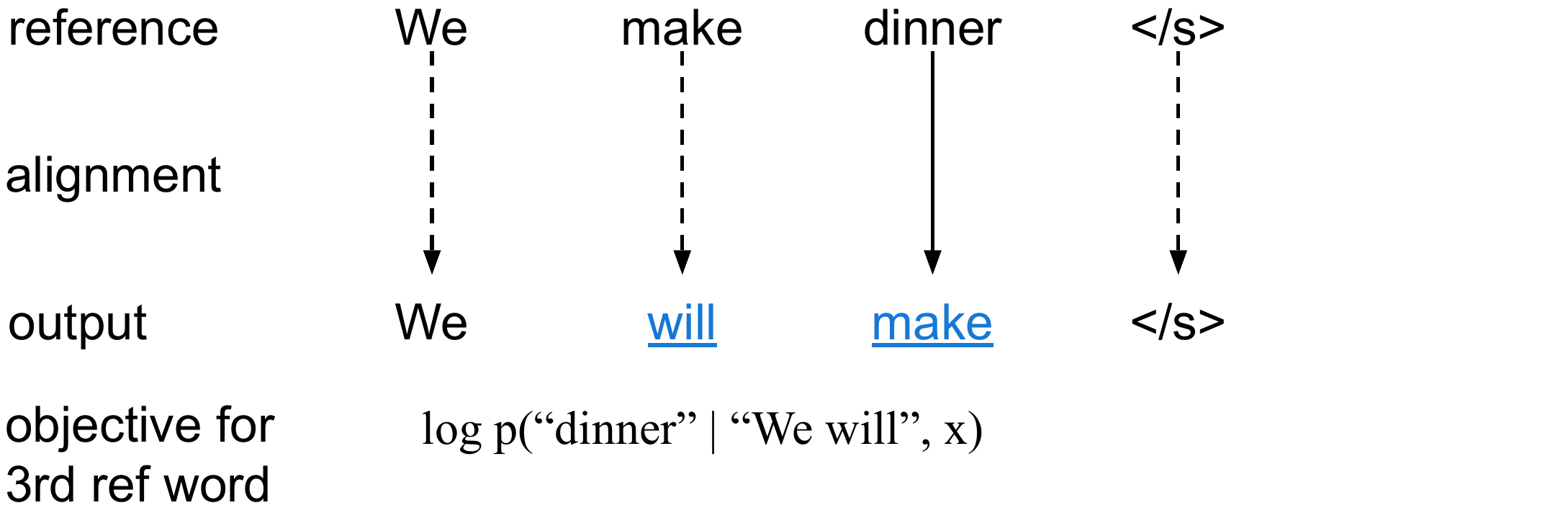} }}
    \hspace*{-3.2em}
    \subfloat[SAML Objective]{{\includegraphics[width=.55\linewidth]{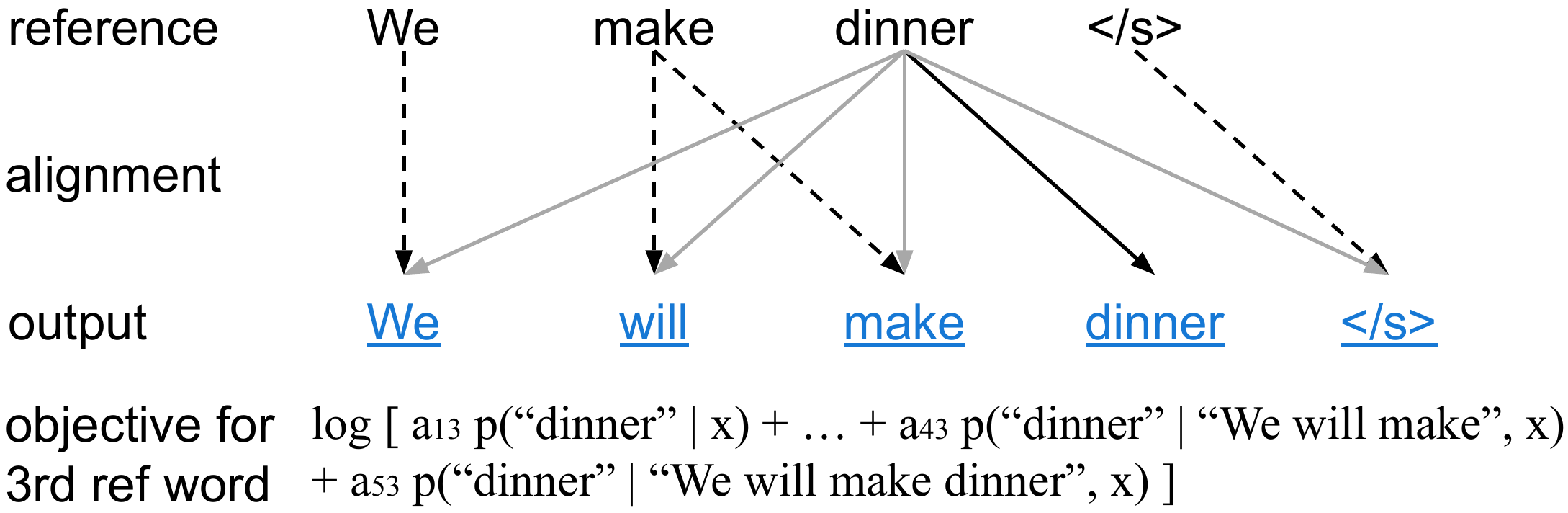} }}
    \caption{Difference between objectives used in scheduled sampling (left) and our approach (right), when computing the contribution to the objective of the reference word ``dinner''. The schedule sampling hypothesis uses a mixture of the reference (black) and sampled (blue underlined) words, while the entire hypothesis sequence is sampled in our approach.}
    \label{fig:example}
\end{figure*}

Our approach is designed to optimize the standard sequence-to-sequence model for translating a source sentence~$\boldsymbol{x}$ into a target sentence~$\boldsymbol{y}$~\cite{BahdanauCB15}. This model computes the probability of~$\boldsymbol{y}$ given~$\boldsymbol{x}$ as:
\begin{equation}
P(\boldsymbol{y}\, | \,\boldsymbol{x}) = \prod_{t=1}^T p(y_t\, | \,\boldsymbol{y}_{<t}, \boldsymbol{x}; \theta)
\end{equation}
where $\theta$ represents the model parameters.
Given~$\boldsymbol{x}$, the model first produces a sequence of hidden representations~$\boldsymbol{h}_{1...T}$:
$\boldsymbol{h}_t = f(\boldsymbol{y}_{<t}, \boldsymbol{x})$,
where~$T$ is the length of~$\boldsymbol{y}$, and~$f$ is usually an encoder-decoder network. At each time step~$t$, the hidden representation~$\boldsymbol{h}_t$ is fed to a linear projection layer~$\boldsymbol{s}_t = \boldsymbol{W} \boldsymbol{h}_t + \boldsymbol{b}$ to obtain a vector of scores~$\boldsymbol{s}_t$ over all possible words in the vocabulary~$\mathcal{V}$. Scores are then turned into a conditional probability distribution:
$p(\cdot\, | \,\boldsymbol{y}_{<t}, \boldsymbol{x}; \theta) = \softmax(\boldsymbol{s}_t)$.

The traditional maximum likelihood (ML) objective maximizes the log-likelihood of the training data~$\mathcal{D} \equiv \{ (\boldsymbol{x}^{(n)}, \boldsymbol{y}^{(n)}) \}_{n=1}^N$ consisting of~$N$ pairs of source and target sentences:
\begin{equation}
\label{equation:mle}
\mathcal{J}_{ML}(\theta) = \sum_{n=1}^N \sum_{t=1}^T \log p(y_t^{(n)}\, | \,\boldsymbol{y}_{<t}^{(n)}, \boldsymbol{x}^{(n)}; \theta)
\end{equation}

At test time, prefixes ~$\boldsymbol{y}_{<t}$ are subsequences generated by the model and therefore contain errors. By contrast, in ML training, prefixes~$\boldsymbol{y}_{<t}$ are subsequences of reference translations. As a result, the model is never exposed to its own errors during training and errors accumulate at test time. This mismatch is known as the exposure bias problem \cite{RanzatoCAZ15}.

\subsection{Limitations in Scheduled Sampling}

\newcite{BengioVJS15} introduced the scheduled sampling algorithm to address exposure bias. Scheduled sampling gradually replaces the reference words with sampled model predictions in the prefix used at training time. An annealing schedule controls the probability of using reference words vs. model predictions. The training objective remains the same as the ML objective, except for the nature of the prefix~$\boldsymbol{\hat{y}}_{<t}$, which contains a mixture of reference and predicted words:
\begin{equation}
\label{equation:ss}
\mathcal{J}_{SS}(\theta) = \sum_{n=1}^N \sum_{t=1}^T \log p(y_t^{(n)}\, | \,\boldsymbol{\hat{y}}_{<t}^{(n)}, \boldsymbol{x}^{(n)}; \theta)
\end{equation}

Despite the empirical success of scheduled sampling, one limitation is that the discontinuity of the argmax operation makes it impossible to penalize errors made in previous steps, which can lead to slow and unstable training. We address this issue using a continuous relaxation to the greedy search and sampling process, similarly to \newcite{GoyalDB17}, which we describe in Section~\ref{sec:sampling}.

Another limitation of scheduled sampling is that it incorrectly assumes that the reference and predicted sequence are aligned by time indices which introduces additional noise to the training signal.\footnote{https://nlpers.blogspot.com/2016/03/a-dagger-by-any-other-name-scheduled.html} We address this problem with a novel differentiable sampling algorithm with an alignment based objective called soft aligned maximum likelihood (SAML). It is used in combination with maximum likelihood to define our training objective~$\mathcal{J} = \mathcal{J}_{ML} + \mathcal{J}_{SAML}$, where~$\mathcal{J}_{ML}$ is computed based on reference translations, and~$\mathcal{J}_{SAML}$ is computed based on sampled translations of the same input sentences. We define~$\mathcal{J}_{SAML}$ in Section~\ref{sec:saml}.

\subsection{Differentiable Sampling}
\label{sec:sampling}

To backpropagate errors made in the previous decoding steps, we use a continuous relaxation of the discrete sampling operation similar to~\newcite{GoyalDB17},
except that we use the Straight-Through (ST) Gumbel-Softmax estimator~\citep{JangGP17,BengioLC13} instead of Gumbel-Softmax~\citep{JangGP17,MaddisonTM14} to better simulate the scenario at inference time.\footnote{The  Straight-Through estimator consistently outperforms the Gumbel-Softmax in preliminary experiments.}

The Gumbel-Softmax is derived from the Gumbel-Max trick~\citep{MaddisonTM14}, an algorithm for sampling one-hot vector~$z \in \mathbb{R}^k$ from a categorical distribution~$(p_1, ..., p_k)$:
\begin{equation}
z = \text{one-hot}(\argmax_i (\log p_i + \beta g_i))
\end{equation}
where~$g_i$ is the Gumbel noise drawn i.i.d from~$\text{Gumbel}(0, 1)$\footnote{$g_i=-\log(-\log(u_i))$ and~$u_i\sim \uniform(0,1)$.}, and~$\beta$ is a hyperparameter controlling the scale of the noise. Here, the trick is used to approximate the discontinuous argmax function with the differentiable softmax:
\begin{equation}
\tilde{z} = \softmax((\log p_i + \beta g_i) / \tau)
\end{equation}
where~$\tau$ is the temperature parameter. As~$\tau$ diminishes to zero,~$\tilde{z}$ becomes the same as one-hot sample~$z$.

The Straight-Through Gumbel-Softmax maintains the differentiability of the Gumbel-Softmax estimator while allowing for discrete sampling by taking different paths in the forward and backward pass. It uses argmax to get the one-hot sample~$z$ in the forward pass, but uses its continuous approximation~$\tilde{z}$ in the backward pass. While ST estimators are biased, they have been shown to work well in latent tree learning~\citep{ChoiYL18} and semi-supervised machine translation~\citep{NiuXC18}.

\subsection{Soft Aligned Maximum Likelihood}
\label{sec:saml}

The soft aligned maximum likelihood (SAML) is defined as  the probability that the reference can be aligned with the sampled output using a soft alignment predicted by the model:
\begin{equation}
\begin{aligned}
P_{SAML}(\boldsymbol{y}\, | \,\boldsymbol{x}) =
\prod_{t=1}^T \sum_{j=1}^{T'} a_{tj} \cdot p(y_t\, | \,\boldsymbol{\tilde{y}}_{<j}, \boldsymbol{x}; \theta)
\end{aligned}
\end{equation}
where~$T$ is the length of the reference sequence,~$T'$ is the length of the sampled sequence,~$a_{tj}$ is the predicted soft alignment between the reference word~$y_t$ and sampled prefix~$\boldsymbol{\tilde{y}}_{<j}$.

Training with the SAML objective consists in maximizing:
\begin{equation}
\mathcal{J}_{SAML}(\theta) = \sum_{n=1}^N \log P_{SAML}(\boldsymbol{y}^{(n)}\, | \,\boldsymbol{x}^{(n)})
\label{equation:saml}
\end{equation}

The conditional probability of the next word~$p(y_t\, | \,\boldsymbol{\tilde{y}}_{<j}, \boldsymbol{x}; \theta)$ is computed as follows:
\begin{equation}
p(\cdot\, | \,\boldsymbol{\tilde{y}}_{<j}, \boldsymbol{x}; \theta) = \softmax(\boldsymbol{W} \boldsymbol{\tilde{h}}_j + \boldsymbol{b})
\end{equation}
where~$\boldsymbol{W}$ and~$\boldsymbol{b}$ are model parameters. $\boldsymbol{\tilde{h}}_j$ is the hidden representation at step~$j$ conditioned on the source sequence~$\boldsymbol{x}$ and the preceding words~$\boldsymbol{\tilde{y}}_{<j}$ sampled from the model distribution using differentiable sampling:
\begin{equation}
\begin{aligned}
\boldsymbol{\tilde{h}}_j = f(\boldsymbol{\tilde{y}}_{<j}, \boldsymbol{x})
\end{aligned}
\end{equation}

We compute the soft alignment~$a_{tj}$ between~$y_t$ and~$\boldsymbol{\tilde{y}}_{<j}$ based on the model's hidden states:
\begin{equation}
a_{tj} =
\frac{ \exp( score(\boldsymbol{\tilde{h}}_j, \boldsymbol{e}_{y_t}) ) }{ \sum_{i=1}^{T'} \exp( score(\boldsymbol{\tilde{h}}_i, \boldsymbol{e}_{y_t}) ) }
\end{equation}
where~$\boldsymbol{e}_{y_t}$ is the embedding of the reference word~$y_t$. The~$score$ function captures the similarity between the hidden state~$\boldsymbol{\tilde{h}}_j$ and the embedding~$\boldsymbol{e}_{y_t}$. We use the dot product here as it does not introduce additional parameters:
\begin{equation}
score(\boldsymbol{h}, \boldsymbol{e}) = \boldsymbol{h}^\top \boldsymbol{e}
\end{equation}

Figure~\ref{fig:example} illustrates how the resulting objective differs from scheduled sampling: (1) it is computed over sampled sequences as opposed to sequences that contain a mixture of sampled and reference words, and (2) each reference word is soft-aligned to the sampled sequence.

%% file: experiments.tex
\begin{table}
\centering
\begin{tabular}{lccccc}
\toprule
\multirow{2}{*}{{\bf Task}} & \multicolumn{3}{c}{sentences (K)} & \multicolumn{2}{c}{vocab (K)} \\
\cmidrule(lr){2-4}
\cmidrule(lr){5-6}
{} & {train} & {dev} & {test} & {src} & {tgt} \\\hline
{\bf de-en} & {153.3} & {7.0} & {6.8} & {113.5} & {53.3} \\
{\bf vi-en} & {121.3} & {1.5} & {1.3} & {23.9} & {50.0} \\\hline
\end{tabular}
\caption{We evaluate on two translation tasks.}
\label{tab:stats}
\end{table}

\begin{table}
\centering
\tabcolsep=0.066cm
\begin{tabular}{lcccc}
\toprule
{\bf Method} & {\bf Anneal} & {\bf de-en} & {\bf en-de} & {\bf vi-en} \\
\midrule
{Baseline} & {No} & {27.41\tiny{$\pm$0.26}} & {22.64\tiny{$\pm$0.13}} & {23.59\tiny{$\pm$0.13}} \\
\midrule
{+SS} & {Yes} & {27.47\tiny{$\pm$0.28}} & {22.56\tiny{$\pm$0.17}} & {23.97\tiny{$\pm$0.39}} \\
{+DSS} & {Yes} & {27.30\tiny{$\pm$0.24}} & {22.47\tiny{$\pm$0.20}} & {23.68\tiny{$\pm$0.35}} \\
{+SS} & {No} & {22.91\tiny{$\pm$0.21}} & {17.78\tiny{$\pm$0.20}} & {19.57\tiny{$\pm$0.19}} \\
\midrule
{+SAML} & {No} & {{\bf 27.94}\tiny{$\pm$0.12}} & {{\bf 23.30}\tiny{$\pm$0.19}} & {{\bf 24.60}\tiny{$\pm$0.35}} \\
\bottomrule
\end{tabular}
\caption{BLEU scores of our approach (SAML) and three baselines including the maximum likelihood (ML) baseline, scheduled sampling (SS), and differentiable scheduled sampling (DSS). The {\it Anneal} column indicates whether the sampling rate is annealed. For each task, we report the mean and standard deviation over 5 runs with different random seeds. SAML achieves the best BLEU scores and is simpler to train than SS and DSS, as it requires no annealing schedule.}
\label{tab:results}
\end{table}

\section{Experiments}

\paragraph{Data} We evaluate our approach on IWSLT 2014 German-English (de-en) as prior work~\citep{GoyalDB17}, as well as two additional tasks: IWSLT 2014 English-German (en-de) and IWSLT 2015 Vietnamese-English (vi-en).
For de-en and en-de, we follow the preprocessing steps in \newcite{RanzatoCAZ15}. For vi-en, we use the data preprocessed by \newcite{StanfordIWSLT15}, with test2012 for validation and test2013 for testing. Table~\ref{tab:stats} summarizes the data statistics.

\paragraph{Setup} Our translation models are attentional RNNs~\citep{BahdanauCB15} built on Sockeye~\citep{CoRR:Sockeye}. We use bi-directional LSTM encoder and single-layer LSTM decoder with~256 hidden units, embeddings of size~256, and multilayer perceptron attention with a layer size of~256. We apply layer normalization~\citep{BaKH16} and label smoothing~(0.1). We add dropout to embeddings~(0.1) and decoder hidden states~(0.2). For ST Gumbel-Softmax, we use temperature~$\gamma = 1$ and noise scale~$\beta = 0.5$. The decoding beam size is~5 unless stated otherwise. We train the models using the Adam optimizer~\citep{KingmaB15} with a batch size of~1024 words. We checkpoint models every~1000 updates.  The initial learning rate is~0.0002, and it is reduced by~30\% after~4 checkpoints without validation perplexity improvement. Training stops after~12 checkpoints without improvement.
\looseness=-1
For training efficiency, we first pre-train a baseline model for each task using only~$\mathcal{J}_{ML}$ and  fine-tune it using different approaches. In the fine-tuning phase, we inherit all settings except that we initialize the learning rate to~0.00002 and set the minimum number of checkpoints before early stopping to~24. We fine-tune each randomly seeded model independently.

\paragraph{Baselines} We compare our model against three baselines:~(1) a standard baseline trained with the ML objective, and models fine-tuned with~(2) scheduled sampling (\textbf{SS})~\citep{BengioVJS15} and~(3) differentiable scheduled sampling (\textbf{DSS})~\citep{GoyalDB17}. In SS and DSS, the probability of using reference words~$\epsilon_s$ is annealed using inverse sigmoid decay~\citep{BengioVJS15}:~$\epsilon_s = k / (k + \exp(i/k))$ at the~$i$-th checkpoint with~$k=10$.

\paragraph{Results} Table~\ref{tab:results} shows that the SAML improves over the ML baseline by +0.5 BLEU on de-en, +0.7 BLEU on en-de, and +1.0 BLEU on vi-en task. In addition, SAML consistently improves over both the scheduled sampling and differentiable scheduled sampling on all tasks. All improvements are significant with~$p < 0.002$. Interestingly, differentiable scheduled sampling performs no better than scheduled sampling in our experiments, unlike in~\newcite{GoyalDB17}.

\begin{figure*}[!t]
	\centering
    \subfloat[de-en]{{\includegraphics[width=0.3\textwidth,height=2.8cm]{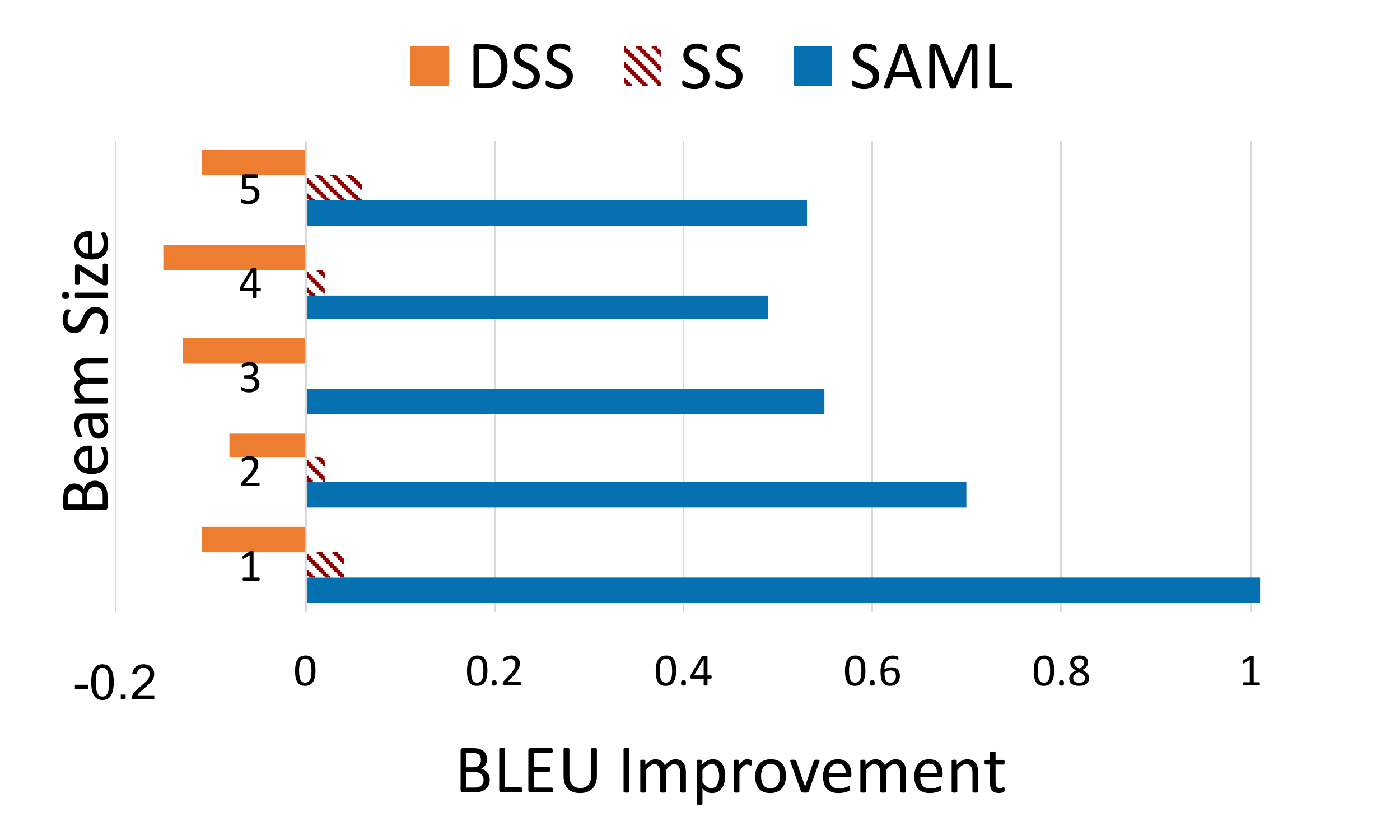} }}\hspace{1em}
    \subfloat[en-de]{{\includegraphics[width=0.3\textwidth,height=2.8cm]{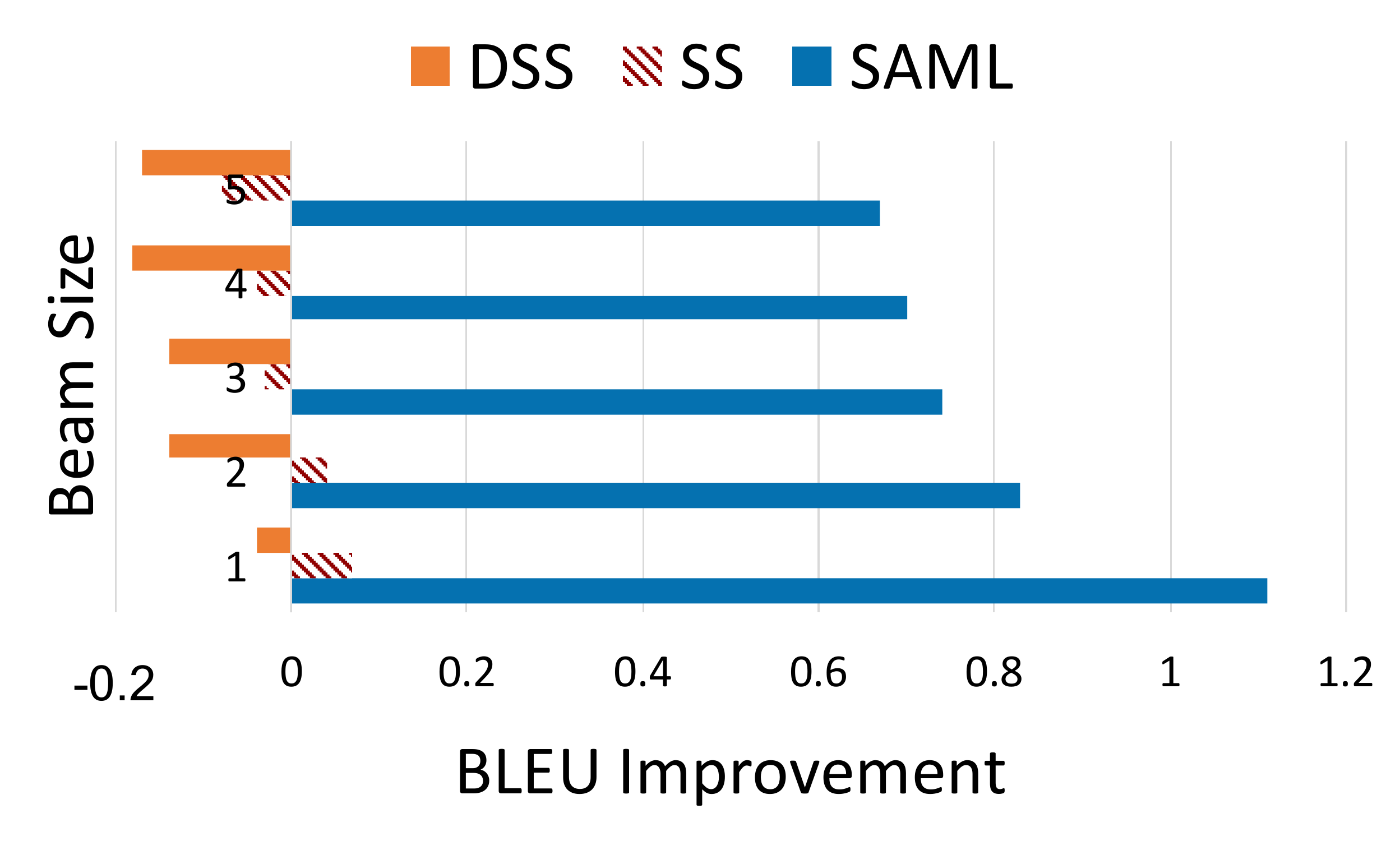} }}\hspace{1em}
    \subfloat[vi-en]{{\includegraphics[width=0.3\textwidth,height=2.8cm]{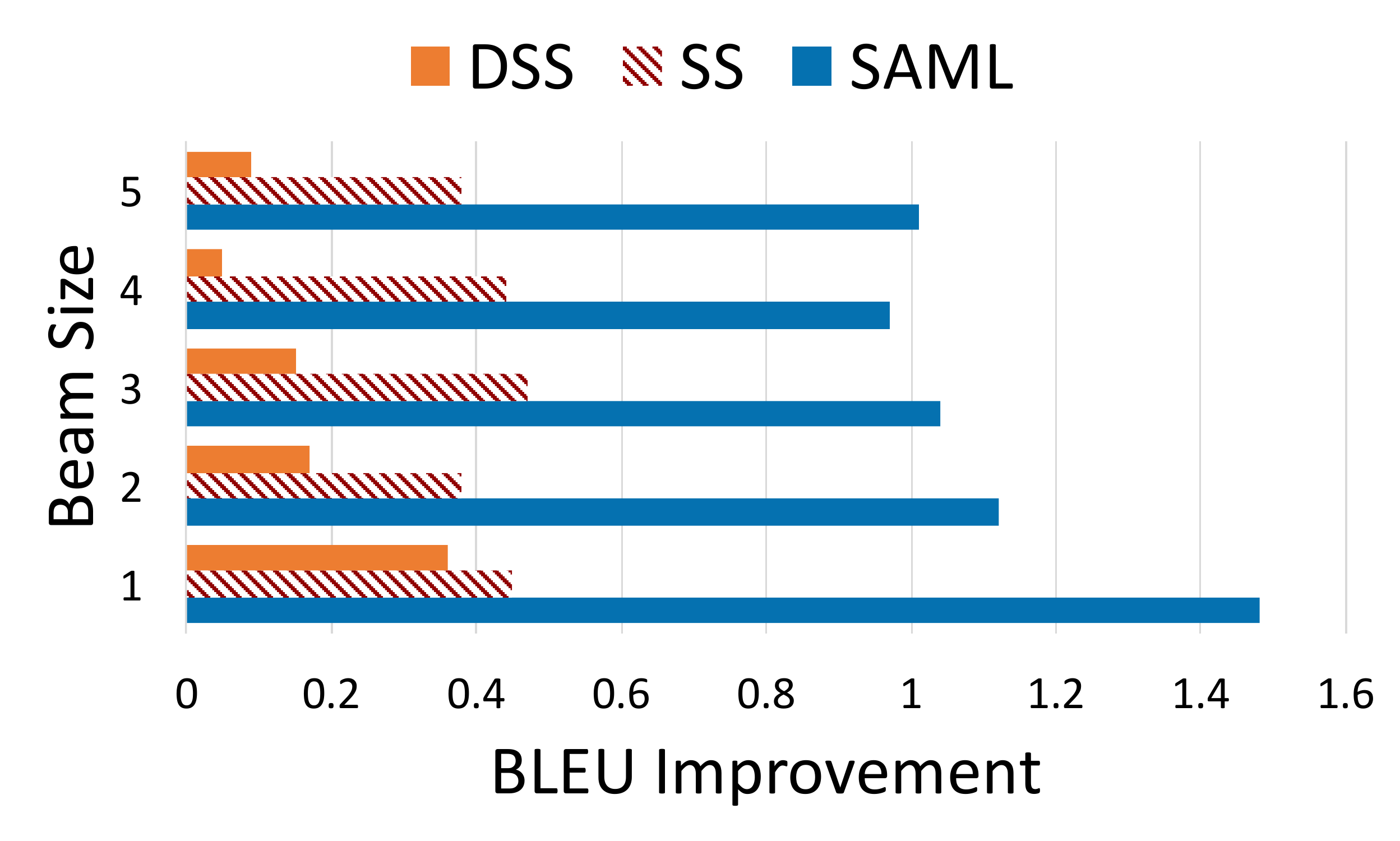} }}
\caption{Improvements from our method (SAML), scheduled sampling (SS), and differentiable scheduled sampling (DSS) over the maximum likelihood (ML) baseline when decoding with varying beam sizes (average of 5 runs). The SAML model consistently yields the largest improvements with smaller beams.}
	\label{fig:delta}
\end{figure*}

Unlike scheduled sampling, our approach does not require an annealing schedule, and it is therefore simpler to train. We verify that the annealing schedule is needed in scheduled sampling by training a contrastive model with the same objective as scheduled sampling, but without annealing schedule (Table~\ref{tab:results}). We set the sampling rate to 0.5. The contrastive model hurts BLEU scores by at least 4.0 points compared to both the ML baseline and models fine-tuned with scheduled sampling, confirming that scheduled sampling needs the annealing schedule to work well.

We further examine the performance gain of different approaches over the baseline with varying beam sizes (Figure~\ref{fig:delta}).
 Our approach yields larger BLEU improvements when decoding with greedy search and smaller beams, while there is no clear pattern for scheduled sampling models. These results support the hypothesis that our approach mitigates exposure bias, as it yields bigger improvements in settings where systems have fewer opportunities to recover from early errors.

%% file: related.tex
\section{Related Work}

\newcite{DaumeLM09} first addressed exposure bias in an imitation learning framework by training a classifier on examples generated using a mixture of the ground truth and the model's current predictions. DAgger~\citep{RossGB11} is a similar algorithm which differs in how the training examples are generated and aggregated. Both algorithms require an expert policy, which produces the best next token given any model predicted prefix, and assume that policy can be efficiently computed from the reference.
However, for structured prediction tasks such as machine translation with large vocabulary and complex loss functions, it is intractable to find the best next token given any prefix.  
For time series modeling, the Data as Demonstrator algorithm~\citep{VenkatramanHB15} derives the expert policy directly from the reference sequences which are aligned with the sampled sequences at each time step. Scheduled sampling algorithms~\citep{BengioVJS15,GoyalDB17} use the same strategy to train neural sequence-to-sequence models for a broader range of language generation tasks, even though the time alignment between reference and sampled sequences does not hold.
\newcite{LeblondAOL18} proposed to complete a predicted prefix with all possible reference suffixes and picking the reference suffix that yields the highest BLEU-1 score. However, they found that this approach performs well only when the prefix is close to the reference.

Reinforcement learning (RL) algorithms~\citep{BahdanauBXGLPCB16,SuttonB18,VanGS16} address exposure bias by directly optimizing a sentence-level reward for the model generated sequences. Evaluation metrics such as BLEU can be used as rewards, but they are discontinuous and hard to optimize.
Techniques such as policy gradient~\citep{Williams1992} and actor-critic~\citep{SuttonB18,DegrisPS12} are thus required to find an unbiased estimation of the gradient to optimize the model. Due to the high variance of the gradient estimation, training with RL can be slow and unstable~\citep{HendersonIBPPM18,WuTQLL18}.
Recent alternatives use data augmentation to incorporate the sentence-level reward into the training objective more efficiently~\cite{NorouziBCJSWS16}.

Finally, our SAML loss shares the idea of flexible reference word order with the bag-of-word loss introduced by \newcite{MaSWL18} to improve source coverage. However, their loss is computed with teacher forcing and therefore does not address exposure bias.